\documentclass[10pt,twocolumn,letterpaper]{article}

\usepackage{cvpr}

\usepackage{xcolor}
\usepackage{soul}

\usepackage[normalem]{ulem}
\usepackage{adjustbox,makecell}
\usepackage{float}
\usepackage{placeins}

\definecolor{darkgreen}{RGB}{0,100,0}

\definecolor{tabfirst}{rgb}{0.96, 0.77, 0.77} 
\definecolor{tabsecond}{rgb}{0.98 , 0.93, 0.77} 
\definecolor{tabthird}{rgb}{1, 1, 0.7} 
\definecolor{lime}{rgb}{0.75, 1.0, 0.0}
\definecolor{theirs}{HTML}{FDAE61}
\definecolor{ours}{HTML}{2B83BA}

\newcommand{\firstplace}[1]{{\textbf{#1}}}
\newcommand{\secondplace}[1]{{\underline{#1}}}

\usepackage{booktabs}      
\usepackage{siunitx}       
\usepackage{adjustbox}     
\usepackage{xcolor}        

\usepackage{algorithm}
\usepackage{listings}

\usepackage{algorithm}
\usepackage{listings}
\usepackage{xcolor}

\definecolor{codegreen}{rgb}{0,0.6,0}
\definecolor{codegray}{rgb}{0.5,0.5,0.5}
\definecolor{codepurple}{rgb}{0.58,0,0.82}
\definecolor{backcolour}{rgb}{0.95,0.95,0.92}

\definecolor{cvprblue}{rgb}{0.21,0.49,0.74}
\usepackage[pagebackref,breaklinks,colorlinks,allcolors=cvprblue]{hyperref}

\title{In-Context Sync-LoRA for Portrait Video Editing}

\author{
Sagi Polaczek$^{1}$ \qquad
Or Patashnik$^{1}$ \qquad
Ali Mahdavi-Amiri$^{2}$ \qquad
Daniel Cohen-Or$^{1}$ \\
{\normalsize
$^{1}$Tel Aviv University \qquad
$^{2}$Simon Fraser University
}
}

\begin{document}
\twocolumn[{%
\renewcommand\twocolumn[1][]{#1}%
\maketitle
\begin{center}
\centering
\captionsetup{type=figure}
\includegraphics[width=\textwidth]{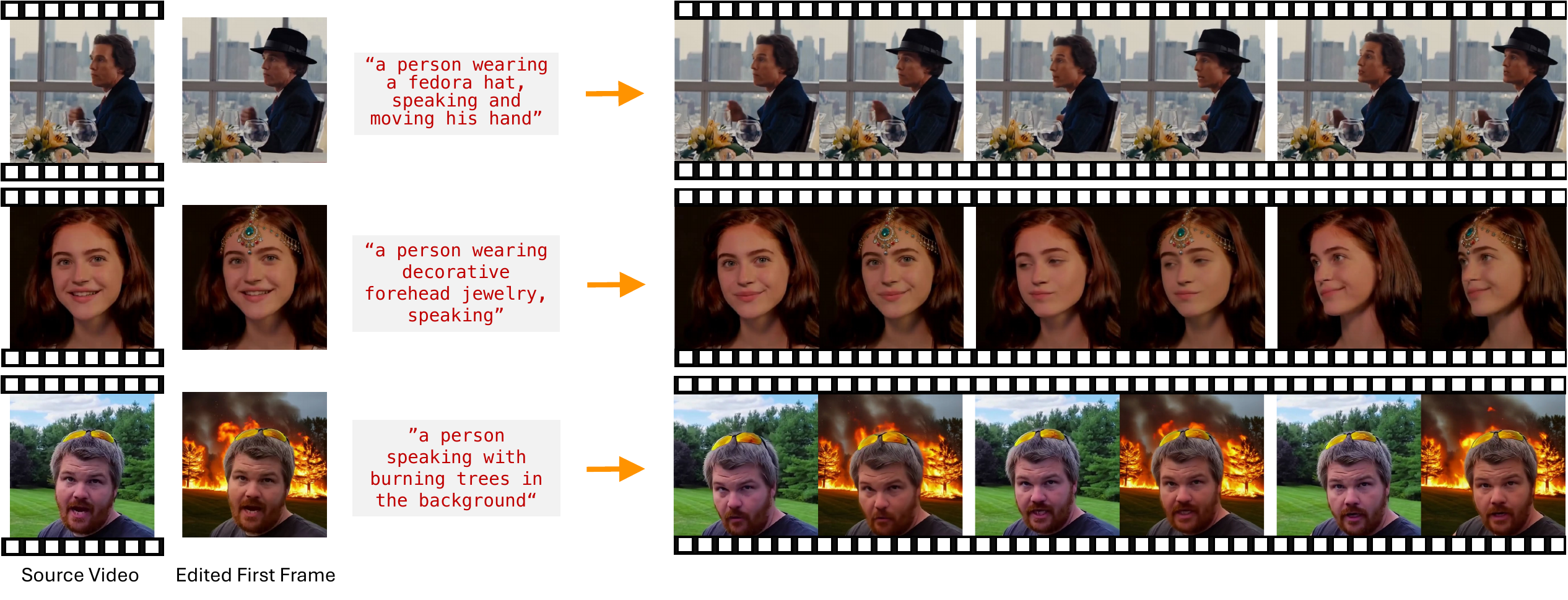}
\caption{Given a source video and an edited first frame, our method propagates the visual edit. The resulting video precisely retains the subject's identity and frame-accurate source motion, while faithfully adopting the new appearance from the edited frame.}
\label{fig:teaser}
\end{center}
}]

\begin{abstract}
Editing portrait videos is a challenging task that requires flexible yet precise control over a wide range of modifications, such as appearance changes, expression edits, or the addition of objects. The key difficulty lies in preserving the subject's original temporal behavior, demanding that every edited frame remains precisely synchronized with the corresponding source frame. We present Sync-LoRA, a method for editing portrait videos that achieves high-quality visual modifications while maintaining frame-accurate synchronization  and identity consistency. Our approach uses an image-to-video diffusion model, where the edit is defined by modifying the first frame and then propagated to the entire sequence. To enable accurate synchronization, we train an in-context LoRA using paired videos that depict identical motion trajectories but differ in appearance. These pairs are automatically generated and curated through a synchronization-based filtering process that selects only the most temporally aligned examples for training. This training setup teaches the model to combine motion cues from the source video with the visual changes introduced in the edited first frame. Trained on a compact, highly curated set of synchronized human portraits, Sync-LoRA generalizes to unseen identities and diverse edits (e.g., modifying appearance, adding objects, or changing backgrounds), robustly handling variations in pose and expression. Our results demonstrate high visual fidelity and strong temporal coherence, achieving a robust balance between edit fidelity and precise motion preservation.
Project page: https://sagipolaczek.github.io/Sync-LoRA/ .
\end{abstract}
    
\vspace{0.0cm}
\section{Introduction}

With the rapid advancement of diffusion-based video editing techniques, text- and image-driven methods have gained prominence for achieving high-quality, temporally consistent edits~\cite{tokenflow2023, vace, wu2023tune}. These techniques hold strong promise for applications in advertising, film production, game development, and interactive media, where fine-grained control over visual content is essential.

A particular interest in video editing lies in the editing of portrait videos, as a large portion of visual media features humans~\cite{guo2024liveportrait, ma2024followyouremoji, Gao2024PortraitGen}. 
Editing portrait videos presents a uniquely demanding challenge.
On the one hand, the visual appearance of the subject must be modified according to a user-defined instruction. On the other hand, the resulting video is expected to preserve the exact motion patterns of the source video.
This requirement goes beyond visual consistency across frames.
The edited video should follow the source on a frame-by-frame basis so that every blink, gaze shift, or articulation occurs at the same moment as in the source.
This level of temporal synchronization is critical in talking-head scenarios, where even small misalignment may render the output unfaithful to the source performance.
At the same time, the desired edit is often localized and specific, such as adding a mustache, changing an expression, or inserting a visual accessory.
The goal is to alter only what is strictly required by the instruction, leaving all other visual content and motion untouched.
Achieving this combination of precise synchronization, a minimal editing footprint, and temporal identity preservation remains an open challenge for most existing video editing techniques~\cite{Gao2024PortraitGen, ku2024anyv2v, molad2023dreamix,vace}.

Inspired by the In-Context LoRA (IC-LoRA) paradigm ~\cite{lhhuang2024iclora, zhang2025ICEdit, chen2025edittransferlearningimage, shin2024diptychprompting, abdal2025dynamic, chen2025multi}, we introduce Sync-LoRA, a method designed to generate precisely synchronized edited portrait videos.
The edit is guided by modifying only the first frame of the source video, which, together with a text prompt, defines the visual transformation to be applied.
Our key idea is to fine-tune an image-to-video diffusion model using curated pairs of videos, one showing the original sequence and the other its edited counterpart.
These paired videos are constructed to be frame-accurately synchronized, meaning that corresponding frames exhibit identical motion trajectories while differing only in the aspects altered by the edit.

To obtain such pairs, we develop a two-stage pipeline: generation followed by synchronization-based filtering.
In the generation stage, a vision-language model produces diverse subject prompts and edit instructions, which are used to synthesize matching portrait-image pairs via text-to-image generation and image editing.  
These are then converted into side-by-side talking-head videos by leveraging the in-context generation capabilities of a diffusion image-to-video model, a concept explored in~\cite{fei2025videodiffusiontransformersincontext}.  
However, these naively generated pairs often suffer from temporal misalignments or motion drift.
To address this critical data quality issue, our filtering stage evaluates each pair's temporal alignment using human motion metrics derived from facial and pose landmarks.
We compute synchronization scores across four complementary channels (speech, gaze, blink, and pose) and \emph{retain only the most aligned pairs} for training (see \autoref{fig:method-data}).
By training on our curated set of synchronized examples, the model learns to propagate motion cues from the source to the edited view, applying only localized changes defined by the first frame while maintaining frame-level alignment with the source video (see Figure~\ref{fig:teaser}).

At inference time, the user simply edits the first frame using any image editing tool.  
Sync-LoRA then takes the original video, the edited frame, and an edit prompt as input, synthesizing an output video that adopts the new appearance while remaining perfectly synchronized with the source.

In contrast to image-based IC-LoRA approaches that require separate training for each editing task, our single Sync-LoRA module generalizes across diverse edits and achieves a robust balance between high visual fidelity and precise motion preservation, as demonstrated in our experiments (see ~\autoref{sec:experiments}).

\section {Related Work}
\begin{figure*}[t]
    \centering
    \includegraphics[width=\linewidth]{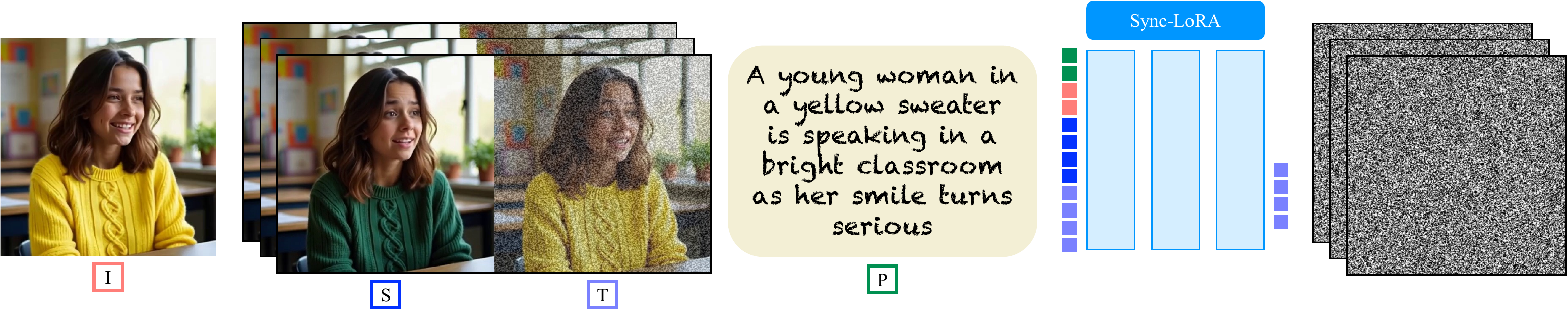}
    \caption{\textbf{Overview of Sync-LoRA.}
    Given a source video $S$, an edited first frame $I$, and an edit prompt $P$, Sync-LoRA denoises the target video $T$ conditioned on these inputs.
    During training, only the edited branch is noised, while the source branch stays clean and provides motion and identity cues through shared attention, so the model copies motion from $S$ and propagates the local edit across all frames.}

    \label{fig:method-training}
\end{figure*}

\paragraph{\textbf{In-Context Learning in Diffusion Models}}
In-context learning (ICL) enables models to generalize to new tasks by conditioning on structured input examples without explicit supervision. While extensively studied in language models~\cite{brown2020language}, recent research has explored ICL in diffusion-based image generation~\cite{lhhuang2024iclora, chen2025edittransferlearningimage, zhang2025ICEdit, shin2024diptychprompting}.
In these works, the model generates a panel of images, where the images within the panel provide context to each other. This enables applications that require consistency across different generated images (e.g., storyboard generation)~\cite{hertz2023StyleAligned, tewel2024consistory, avrahami2024chosen}.
When conditioning such panels on a given real image, ICL-based approaches can be used for a wide range of image-to-image tasks, such as image personalization~\cite{gal2022textual, ruiz2023dreambooth, tan2024ominicontrol, kumari2025syncd, cai2024dsd, gal2024lcmlookahead} and image editing~\cite{Patashnik_2021_ICCV, hertz2022prompt, parmar2023zero, garibi2024renoise, najdenkoska2025contextdiffusionincontextaware}.

The most closely related work to ours is IC-LoRA~\cite{lhhuang2024iclora}. This work observes that transformer-based diffusion models possess an inherent capability to generate panels in which certain attributes remain consistent across the different images in the panel. They show that this capability can be enhanced by fine-tuning the model with a low-rank adapter (LoRA) on structured, consistent panels, and using an appropriate template prompt. Their approach has been applied to diverse tasks such as film storyboard generation, font design, and home decoration. Our work extends the paradigm to video, and presents an IC-LoRA for video editing which focuses on frame-accurate portrait synchronization.

Additionally, recent progress has shown that large video diffusion transformers naturally exhibit in-context learning behavior, enabling them to generate multiple temporally or spatially related clips within a single diffusion process~\cite{fei2025videodiffusiontransformersincontext}. 
That work focuses on analyzing this emergent ability for in-context \emph{generation}, using inpainting-based conditioning to achieve controlled scene composition, but without any mechanism to ensure temporal correspondence across clips. 
Building on this insight, PoseGen~\cite{he2025posegenincontextlorafinetuning} introduces in-context LoRA finetuning for pose-controllable human video generation, achieving long and coherent motion from a reference image. 
In contrast, our approach explicitly \emph{trains} this in-context behavior toward \emph{synchronized editing}.

\paragraph{\textbf{LoRA for Video Diffusion Models}}
Recent works have extended LoRA-based fine-tuning to video diffusion models, primarily focusing on subject or motion personalization. Customize-A-Video~\cite{ren2024customize} applies LoRA to the temporal attention layers of a text-to-video diffusion model, enabling motion customization. CustomTTT~\cite{wu2024customttt} introduces a method for generating customized videos in which the subject is taken from a reference image and the motion from a source video. This work identifies motion- and appearance-specific layers and applies LoRA modules for per-layer adaptation. Abdal et al.~\cite{abdal2025dynamic} introduce the notion of personalizing dynamic concepts. In this task, the model learns both the appearance and motion of a concept from a reference video. They achieve this by training a LoRA on the reference video, which then enables generation of the learned dynamic concept in novel scenes.
While these works focus on personalizing appearance or motion, our method is designed for frame-accurate synchronized editing, which demands far stricter preservation of the source video's motion.

\vspace{-0.4cm}
\paragraph{\textbf{Video Editing}}
Recent research on diffusion-based video editing has introduced methods that use text or image prompts to enable user-guided modifications. Some methods~\cite{decart2025lucyedit, molad2023dreamix, wu2023tune, liu2023videop2p, ceylan2023pix2video} focus on text-based editing, typically excelling at appearance modifications that preserve motion. Others~\cite{ku2024anyv2v, i2vedit,vace, ouyang2023codef} apply edits to the first frame using existing image editing techniques, and then propagate the changes across the video. Additional methods have been developed for specific tasks such as object insertion~\cite{yatim2025dynvfx} or overlaying hand-drawn doodles onto videos~\cite{videodoodles}.

In portrait video editing, earlier approaches rely on keypoints \cite{FOMM,Wang2021Oneshot,Koyejo2022Implicit,elorportraits,guo2024liveportrait} or 3DMM template meshes~\cite{Khakhulin2022ROME,Tewari2020StyleRig} to transfer expressions from a driving video to a source image. More recently, diffusion models have enabled new methods that generate videos from a single image guided by a driving video. These methods typically establish correspondence either through keypoints~\cite{ma2024followyouremoji} or 3DMM~\cite{Zeng_2023_CVPR} as an intermediate representation or by leveraging the attention mechanisms within diffusion models~\cite{wei2024aniportrait,Xie2024XPortrait,yang2024megactor}.
In contrast, our approach uses in-context learning: the edited first frame sets the appearance context, while the source video provides the motion context, eliminating the need for explicit correspondence modeling.

\section{Method}\label{method}

\subsection{In-Context LoRA}

Inspired by the IC-LoRA paradigm, we develop a framework for portrait video editing. 
Our model learns to generate an edited video conditioned on a clean source video and its edited first frame, which together serve as contextual input.  
This setup enables synchronized, context-aware editing during inference, where conditioning on the input video naturally steers the generation of its edited counterpart.

While IC-LoRA was originally designed for static image panels, extending it to video requires the model to reason not only about appearance correspondence but also about temporal alignment across frames. 
To handle this additional temporal dimension, we build on a transformer-based diffusion model (DiT) designed for image-to-video generation (see \autoref{fig:method-training}). 
In this architecture, the input image (first frame), source video, and text prompt are represented as token sequences, and joint-attention layers operate over them to iteratively denoise the edited video representation. 

During training, only the edited view is denoised, while the source view remains noise-free and provides motion guidance through shared attention. 
This design directs the model to learn temporal synchronization between the two views rather than re-solving the edit, as the appearance change is already defined by the conditioning panel. 
At inference, the user edits the first frame, and the model generates a coherent video that preserves the original motion.

Alongside the visual conditioning, the model is also guided by a text prompt that describes the intended relation between the two views (green tokens in \autoref{fig:method-training}). 
Conditioned on the first frame and text prompt, the model learns to propagate the edit consistently across the entire sequence while maintaining precise motion correspondence.

\subsection{Data Generation and Curation}
\label{subsec:data_generation_and_curation}

\begin{figure}[t]
    \centering
    \includegraphics[width=\linewidth]{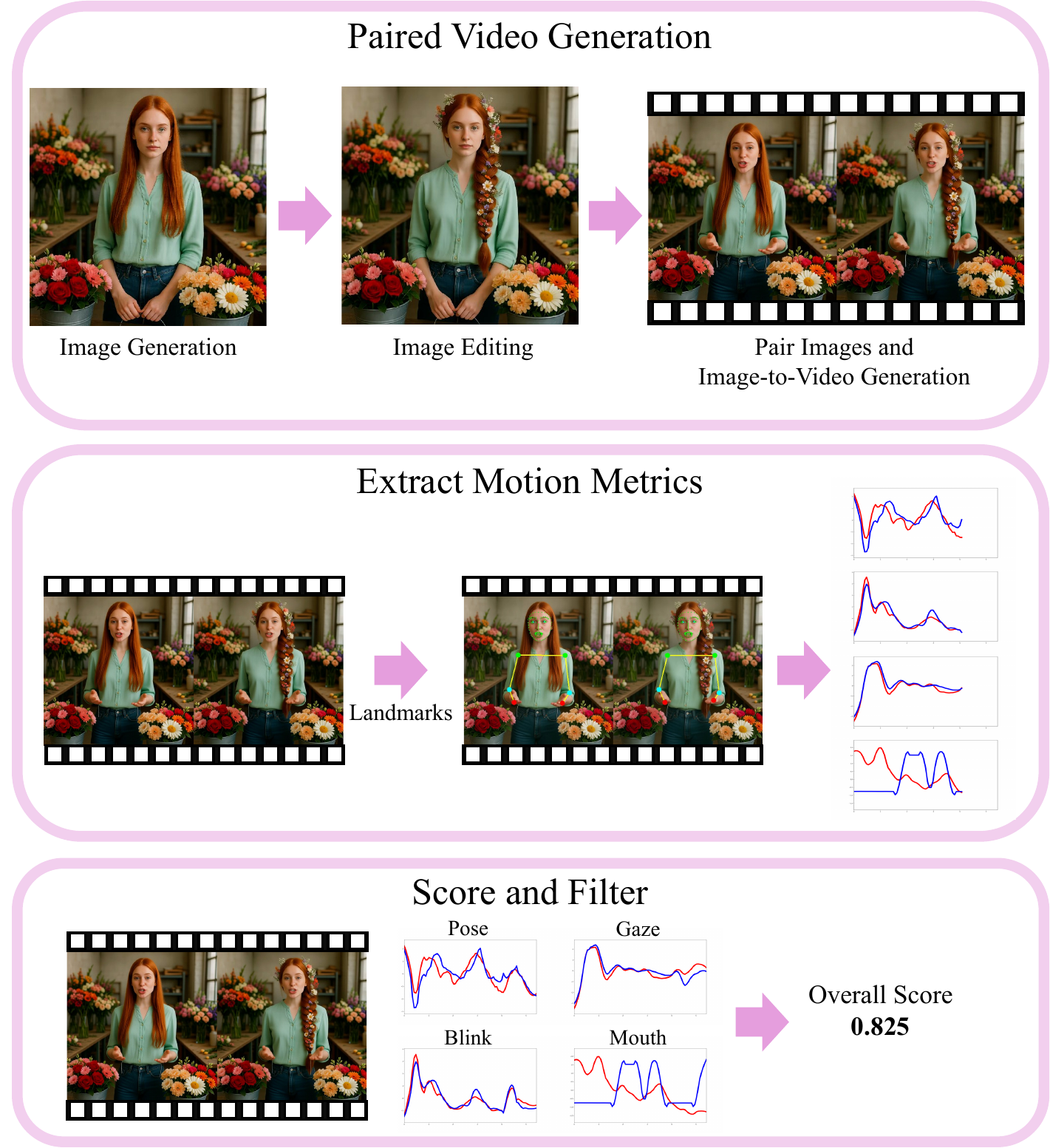}
    \caption{\textbf{Data generation and curation pipeline.}\\
    Our process constructs synchronized video pairs for Sync-LoRA training. 
    \textbf{(Top)} Portrait images are generated, edited, and converted into side-by-side talking-head videos. 
    \textbf{(Middle)} Facial and pose landmarks yield motion signals for speech, gaze, blink, and pose. 
    \textbf{(Bottom)} Pairs are scored and filtered by synchronization quality, keeping only the most aligned examples for training.
    }
    \label{fig:method-data}
\end{figure}

Training Sync-LoRA requires high-quality talking-head video pairs that differ in appearance but exhibit tight temporal synchronization.
These video pairs serve as the foundational supervision signal, and, as shown in our ablations, their precision is critical for faithfully transferring motion dynamics while applying the necessary edits (see \autoref{subsec:ablations}).
To create such data, we devise a two-stage pipeline: large-scale generation followed by synchronization-based filtering.
The full process is illustrated in \autoref{fig:method-data}.

\vspace{-0.4cm}
\paragraph{\textbf{Generation.}}  
We generate a diverse, large-scale synthetic talking-head pairs.  
A vision-language model (VLM) produces subject prompts and edit instructions (e.g., “change hair color,” “add a hat”), to create aligned portrait-image pairs via image generation and editing.  
Each pair is composed into a dual-panel image and converted into a side-by-side talking-head video using an image-to-video model.  
The process leverages the in-context generation ability of our base \emph{data-generation} model (Wan2.1~\cite{wan2025wan}), which we further refine by LoRA fine-tuning on a small, curated set of synchronized examples to improve motion consistency.

\begin{figure}[t]
    \centering
    \includegraphics[width=\linewidth]{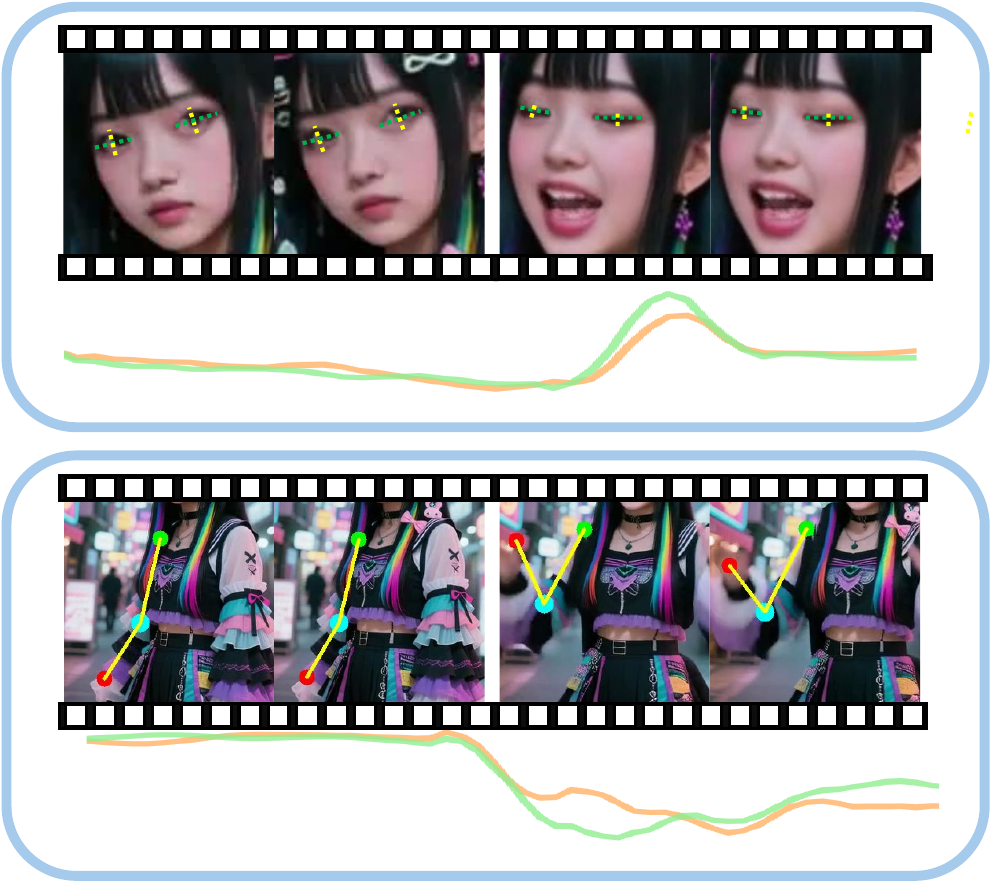}
\caption{\textbf{Synchronization signal visualization.} 
    Two synchronization cues used in our filtering process. 
    \textbf{Top:} Eye landmarks are used to compute the Eye Aspect Ratio (EAR). Note how the plotted peaks (reference in green, edited in orange) correspond directly to the blink event shown in the frames above. 
    \textbf{Bottom:} Upper-body pose landmarks are used to track the right elbow angle. The plots again show tightly correlated motion, confirming the arm movement is synchronized across both videos.}    \label{fig:method-filtering-one-col}
\end{figure}

\vspace{-0.4cm}
\paragraph{\textbf{Filtering.}}

Despite being designed for synchronization, generated pairs often contain misalignments.  
To ensure data quality, we introduce a filtering stage that quantifies temporal correspondence using motion-based metrics derived from facial and pose landmarks.
Human facial and pose landmarks are extracted using MediaPipe~\cite{lugaresi2019mediapipeframeworkbuildingperception}, producing per-frame scalar signals for analysis.
For each pair, trajectories of key landmarks are analyzed across four motion channels ,speech, gaze, blink, and pose, capturing complementary aspects of talking-head dynamics.  
Signals are interpolated to fill missing detections, smoothed with a Savitzky-Golay filter~\cite{Savitzky–Golay}, z-normalized, and correlated between the left and right views at zero temporal lag. Visualization is shown in \autoref{fig:method-filtering-one-col} and full implementation details are provided in the supplementary material.
Pairs with insufficient detection coverage are discarded, and top-ranked examples are retained, yielding high-quality synchronized videos.

\begin{figure*}[!t]
\centering
\includegraphics[width=\textwidth]{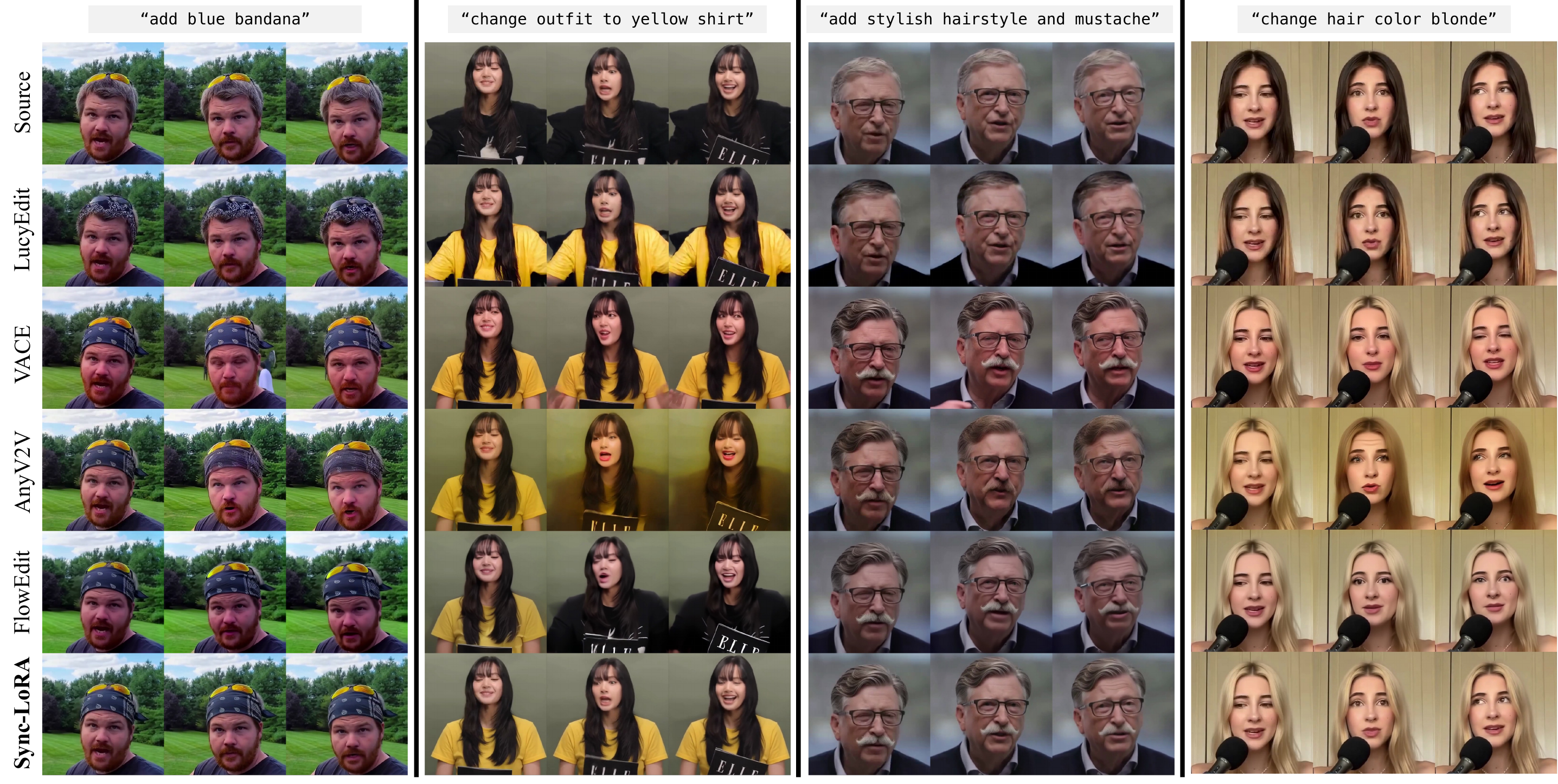}
\caption{\textbf{Comparison of portrait video editing methods.}
The rows show the source video and results from LucyEdit, VACE, AnyV2V, FlowEdit, and Sync-LoRA (Ours).
The columns depict different temporal positions.
Our method, VACE, AnyV2V, and FlowEdit utilize the same edited first frame as visual input, whereas the text-based LucyEdit operates from text guidance alone.
}
\label{fig:comparison}
\end{figure*}

\vspace{-0.4cm}
\paragraph{\textbf{Synchronization Scoring Metrics.}}
We measure \emph{speech} using the mouth aspect ratio over time.
\emph{Blink} uses the Eye Aspect Ratio (EAR)~\cite{cech2016realtimeeyeblinkdetection}, with peaks indicating closure events.
\emph{Gaze} tracks normalized 2D iris motion, while \emph{pose} uses six upper-body angles or relative heights (shoulder, torso, elbows, wrists).
Each signal is independently correlated between the original and edited videos and then combined into a weighted synchronization score (weights: 40\% speech, 30\% gaze, 15\% blink, 15\% pose).
This curated dataset forms the foundation of Sync-LoRA, enabling precise motion preservation while applying localized visual edits.
Full definitions and implementation details are provided in the supplementary material.

\vspace{-0.4cm}
\paragraph{\textbf{Implementation Details.}} 
We use QwenImage and QwenImageEdit-2509~\cite{wu2025qwenimagetechnicalreport} for portrait generation and editing the first frame, respectively.
For video synthesis, we employ Wan2.1 ~\cite{wan2025wan} as the base image-to-video model, generating each paired video with 30 denoising steps.

We generate over 20,000 paired examples and apply a synchronization-based filtering to retain the most temporally aligned samples.  
The final training set consists of only 512 video pairs, with a 3:1 ratio of edited to identical (unmodified) samples to stabilize training.

\subsection{Training}

\paragraph{\textbf{Base Model.}} 
We build our method upon LTX-Video~\cite{hacohen2024ltxvideorealtimevideolatent}, a transformer-based latent diffusion model that integrates the video-VAE and the denoising transformer in a unified framework. 
LTX-Video operates in a highly compressed spatio-temporal latent space and employs full 3D attention, enabling efficient generation of high-resolution, temporally coherent videos. 

To adapt this model for synchronized portrait editing, we fine-tune it using low-rank adaptation (LoRA)~\cite{hu2022lora} with a rank of 128.
This lightweight adaptation preserves the rich prior of LTX-Video while encouraging the model to rely on the provided first frame for appearance and motion cues. 
This setup ensures consistent temporal alignment and visual coherence, without altering the base architecture.
Training is facilitated by the LTX-Video-Trainer code~\cite{LTXVideoTrainer2025}.

\vspace{-0.4cm}
\paragraph{\textbf{Positional Encoding.}}  
Following LTX-Video~\cite{hacohen2024ltxvideorealtimevideolatent},  
each latent token is represented by a continuous 3D coordinate $(t, h, w)$ within the spatio-temporal volume,  
scaled by VAE compression factors and normalized by frame rate.  
We embed these coordinates using 3D Rotary Positional Embeddings (RoPE)~\cite{su2023roformerenhancedtransformerrotary}.  
While both source and target streams share identical positional coordinates for spatial alignment, they are differentiated by \emph{per-token timesteps}.
This mechanism, inherited from the LTX-Video base model~\cite{hacohen2024ltxvideorealtimevideolatent}, uses each token's timestep to generate unique \emph{scale} and \emph{shift} parameters for its Adaptive Layer Normalization (AdaLN).
The model thus learns to treat source stream tokens (assigned $t=0$) as clean conditioning, and target stream tokens (assigned $t>0$) as the noisy sequence to be denoised.

\vspace{-0.4cm}
\paragraph{\textbf{Loss Function.}}  
We adopt the rectified flow objective~\cite{lipman2023flowmatchinggenerativemodeling,esser2024scalingrectifiedflowtransformers}, also used in LTX-Video, to train our model to predict the velocity field that transforms noise into the clean latent of the edited video.  
Unlike conventional diffusion setups generating a single video from text or an image prompt, our formulation introduces an additional conditioning stream, the source video, which remains noise-free throughout denoising and provides motion and appearance cues.

Given a noisy latent sample of the edited branch  
$x_t = (1 - t)\,x_1 + t\,x_0$ with timestep $t \in [0, 1]$,  
the model is optimized to match the ground-truth velocity $v_t = x_t - x_0$:

\vspace{-0.4cm}

\begin{equation*}
    \mathcal{L}_{\text{RF}} 
    = \mathbb{E} 
    \left[ 
        \left\| 
        u(x_t, c_{\text{text}}, c_{\text{img}}, c_{\text{vid}}, t; \theta_{\text{LoRA}}) 
        - v_t 
        \right\|_2^2
    \right],
\end{equation*}

where $x_0 \sim \mathcal{N}(0, I)$ is Gaussian noise,  
$x_1$ is the target latent of the edited video,  
$c_{\text{text}}$ is the editing prompt,  
$c_{\text{img}}$ is the edited first frame,  
and $c_{\text{vid}}$ represents the latent features of the source video.

\section{Experiments}
\label{sec:experiments}

\paragraph{\textbf{Evaluation Setup.}}  
We evaluate Sync-LoRA on a curated benchmark of 166 portrait videos assessing spatial fidelity, and temporal alignment.  
The benchmark spans diverse edit types, including object insertion, background replacement, colorization, and appearance modification, with source clips from CelebV~\cite{yu2023celebvtextlargescalefacialtextvideo}, CelebV-HQ~\cite{zhu2022celebvhqlargescalevideofacial}, TalkVid~\cite{chen2025talkvidlargescalediversifieddataset}, and high-quality YouTube content.  

To ensure consistency, we apply strict curation, retaining only sharp, well-lit clips of at least five seconds. Clips with subtitles, letterboxing, cropping, or jump cuts are excluded. Representative accepted and rejected samples are shown in the supplementary.

For each sequence, the first frame is edited using QwenImageEdit-2509~\cite{wu2025qwenimagetechnicalreport}, guided by VLM-generated instructions to cover a wide range of transformations.  
We compare our method against four strong baselines, VACE~\cite{vace}, LucyEdit~\cite{decart2025lucyedit}, FlowEdit~\cite{kulikov2025floweditinversionfreetextbasedediting}, and AnyV2V~\cite{ku2024anyv2v}, covering both qualitative and quantitative evaluations across state-of-the-art video editing paradigms.
The VACE, LucyEdit, and AnyV2V results are based on their official implementations, and for FlowEdit, we used the public LTX-Video implementation.
All clips contain 81 frames at 20~FPS.

\subsection{Qualitative Evaluation}

\paragraph{\textbf{Qualitative Comparison.}}
As shown in \autoref{fig:comparison}, Sync-LoRA produces temporally coherent and visually faithful results across the diverse editing tasks illustrated in the figure.
While all baselines attempt the edit, they exhibit notable failures.
Using the third column as a representative example, LucyEdit fails to apply the edit faithfully, while VACE, AnyV2V, and FlowEdit all introduce temporal artifacts, inconsistent textures, or motion drift.
In contrast, Sync-LoRA faithfully preserves the source video's motion while consistently applying the edits across all frames.
Our method achieves superior synchronization and visual fidelity in all examples, demonstrating frame-accurate alignment and consistent visual appearance throughout the sequence.
The full temporal performance of our method and all baselines is best assessed in the supplementary video, which highlights Sync-LoRA's ability to maintain precise synchronization while achieving high edit fidelity.

\subsection{Quantitative Comparisons}

We evaluate the generated outputs across three key aspects: \emph{Synchronization}, \emph{Edit Fidelity}, and \emph{Identity Preservation}.

\noindent\textbf{(1) Synchronization.}
We measure frame-level temporal correspondence between the source and edited videos using four correlation-based metrics: \emph{speech}, \emph{gaze}, \emph{blink}, and \emph{pose}.
Each captures a distinct aspect of motion dynamics: mouth aspect ratio for speech, iris trajectory for gaze, eye aspect ratio for blinking, and upper-body joint angles for pose.
While these metrics are identical to those used for our training data filtering (\autoref{subsec:data_generation_and_curation}), they serve only to curate the dataset and perform this final evaluation.
They are never used as a loss function or otherwise included in the model's training objective.

\begin{table}[t]
\small
\setlength{\tabcolsep}{5pt}
\addtolength{\belowcaptionskip}{-6pt}
\centering
\begin{adjustbox}{width=\linewidth}
\begin{tabular}{l S[table-format=0.2] S[table-format=0.2] S[table-format=0.2] c S[table-format=0.2] S[table-format=0.2] S[table-format=0.2]}
    \toprule
    Metric & \multicolumn{3}{c}{VACE} & {LucyEdit} & {FlowEdit} & {AnyV2V} & {\textbf{Ours}} \\
    \cmidrule(r){2-4}
           & {Pose} & {Canny} & {Depth} & & & & \\
    \midrule
    \multicolumn{8}{l}{\textbf{Synchronization}} \\
    Speech Corr. $\uparrow$ & 0.37 & 0.11 & 0.39 & \firstplace{0.80} & 0.50 & 0.70 & \secondplace{0.72} \\
    Gaze Corr. $\uparrow$   & 0.63 & 0.63 & 0.69 & \firstplace{0.82} & 0.56 & 0.71 & \secondplace{0.75} \\
    Blink Corr. $\uparrow$  & 0.34 & 0.24 & 0.36 & \firstplace{0.70} & 0.33 & 0.48 & \secondplace{0.55} \\
    Pose Corr. $\uparrow$   & 0.45 & 0.39 & 0.51 & \firstplace{0.64} & 0.38 & 0.51 & \secondplace{0.55} \\
    \midrule
    \multicolumn{8}{l}{\textbf{Edit Fidelity}} \\
    Directional CLIP (image) $\uparrow$ & \firstplace{0.57} & 0.55 & 0.45 & {N/A} & 0.53 & 0.39 & \firstplace{0.57} \\
    Directional CLIP (text-dual) $\uparrow$ & \secondplace{0.20} & \secondplace{0.20} & 0.16 & 0.11 & 0.18 & 0.17 & \firstplace{0.21} \\
    CLIP-Text Align. $\uparrow$ & \firstplace{0.33} & 0.33 & 0.32 & 0.32 & 0.32 & 0.32 & \firstplace{0.33} \\
    \midrule
    \multicolumn{8}{l}{\textbf{Identity Preservation}} \\
    ArcFace Sim. $\uparrow$ & \secondplace{0.73} & 0.71 & 0.72 & 0.69 & 0.68 & 0.63 & \firstplace{0.75} \\
    \bottomrule
\end{tabular}
\end{adjustbox}
\vspace{0.0cm}
\caption{\textbf{Quantitative Video Comparisons.}
We evaluate across three axes: \emph{Synchronization}, \emph{Edit Fidelity}, and \emph{Identity Preservation}.
The \textbf{Directional CLIP (image)} score is omitted for \emph{LucyEdit}, as it operates solely from text prompts, making this metric not directly applicable.
}
\label{tb:video_metrics}
\end{table}

\noindent\textbf{(2) Edit Fidelity.}  
We evaluate whether the intended edit is accurately applied and remains consistent throughout the video using two complementary CLIP-based metrics.  

\noindent\underline{\textit{Directional CLIP Score.}}
Originally proposed for image editing, we adapt the Directional CLIP metric to the video domain to measure how consistently the per-frame visual transformation aligns with the overall edit direction in CLIP’s embedding space.
We report two complementary variants: an \emph{image-based} version that quantifies alignment with the visual edit implied by the edited first frame, and a \emph{text-dual} version that measures alignment with the semantic edit described by the text prompt.  
The text-dual variant is included for fairness, since some baselines such as LucyEdit operate purely from textual guidance and do not have access to the edited first frame.
Intuitively, the image-based variant asks whether the change from source frames to edited frames looks like the change between the unedited and edited first frames, while the text-dual variant asks whether that same visual change looks like the change suggested by the edit text.

\noindent\underline{\textit{CLIP-Text Alignment Score.}}
To further assess semantic accuracy, we measure how well each generated frame matches the  textual description in CLIP space.  
This score captures whether the video remains faithful to the prompt semantics across time, complementing the directional measure above.
In contrast to the directional score, CLIP-Text Alignment ignores the source frames and judges only how well each edited frame by itself matches the target description.

Together, these metrics jointly evaluate both the \emph{directional consistency} of the applied edit and its \emph{semantic alignment} throughout the sequence.

\noindent\textbf{(3) Identity Preservation.}
To evaluate how well subject identity is maintained throughout the edited video, we compute the mean cosine similarity between face embeddings extracted by a pre-trained ArcFace~\cite{Deng_2022_ArcFace} model for each frame and the edited first frame.  
This measure reflects how faithfully the appearance and facial structure of the person are preserved during temporal propagation, where higher values correspond to stronger identity consistency.  

Furthermore, a user study confirmed our method is perceptually preferred for visual quality and temporal alignment. Full details for this study and all quantitative metric formulations are provided in the supplementary material.

\noindent\textbf{Results Discussion.}  
As shown in ~\autoref{tb:video_metrics}, while text-based methods like LucyEdit achieve high scores on some sync metrics (e.g., Speech Corr. 0.80 vs. our 0.72), this comes at the cost of failing the edit itself, as shown by its low 0.11 Directional CLIP (text-dual) score (vs. our 0.21).
Our method achieves the \textbf{best overall balance}, strongly performing on synchronization while also faithfully applying the visual edit.
VACE, conversely, benefits from strong external conditioning (pose/canny/depth), yields higher CLIP-based alignment, yet it exhibits weaker temporal correspondence and identity consistency.
By explicitly training for synchronized editing, Sync-LoRA maintains strong temporal alignment while producing precise edits that preserve subject identity. 
This trade-off highlights the strength of our in-context formulation, which balances motion coherence and edit fidelity rather than excelling in a single axis.

\begin{figure}[t]
\centering
\begin{adjustbox}{width=\linewidth}
\includegraphics[width=\textwidth]{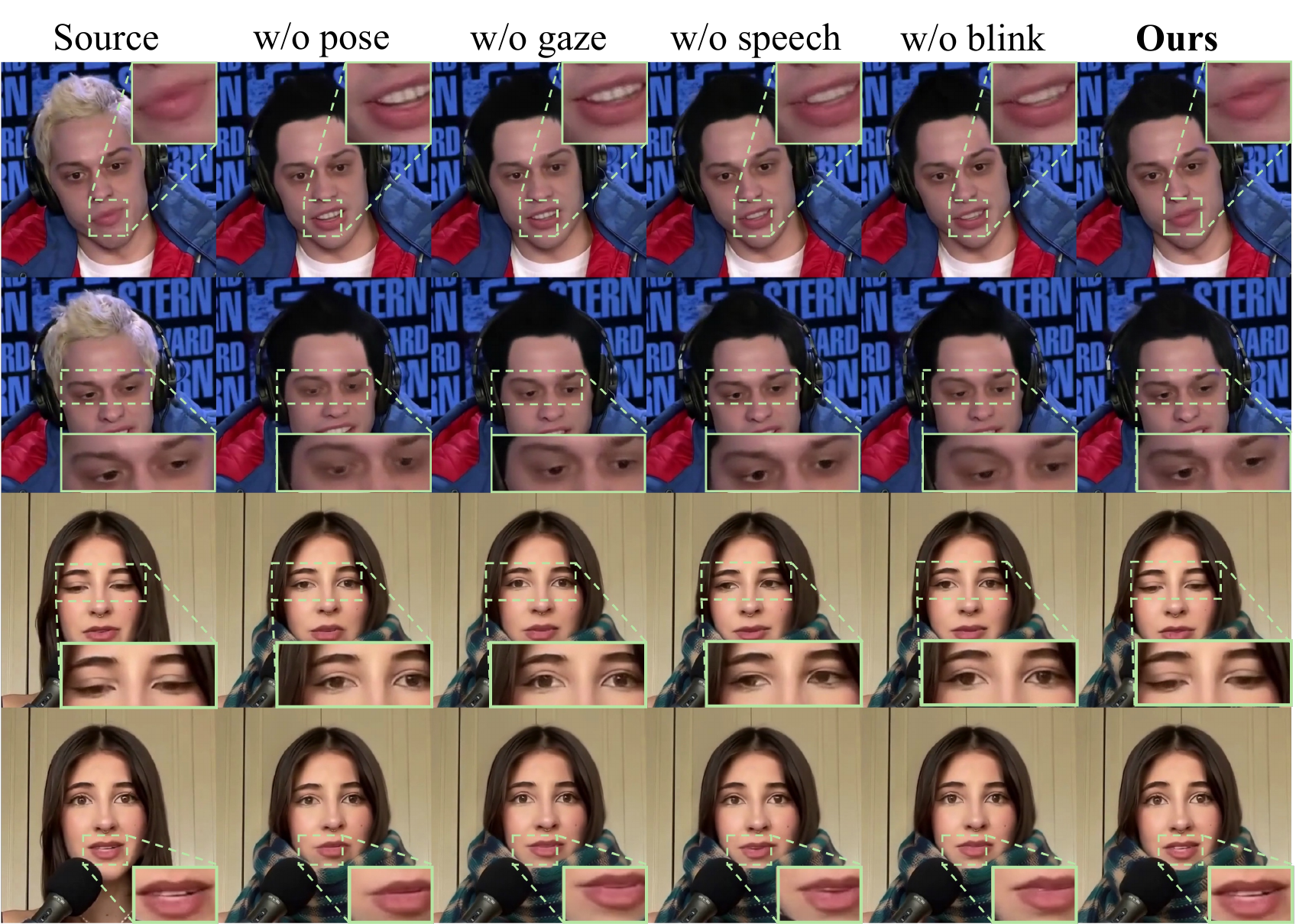}
\end{adjustbox}

\vspace{-0.2cm}

\caption{
\textbf{Necessity of all synchronization cues.} 
Each column shows results when training without the specified motion cue (pose, gaze, speech, or blink) from the filtering stage, compared to our full setup. The source video is shown on the left. Omitting any cue causes motion drift or misalignment across frames.}
\vspace{-0.2cm}

\label{fig:ablation_leave_one_out}
\end{figure}

\begin{figure}[t]
\centering
\begin{adjustbox}{width=\linewidth}
\includegraphics[width=\textwidth]{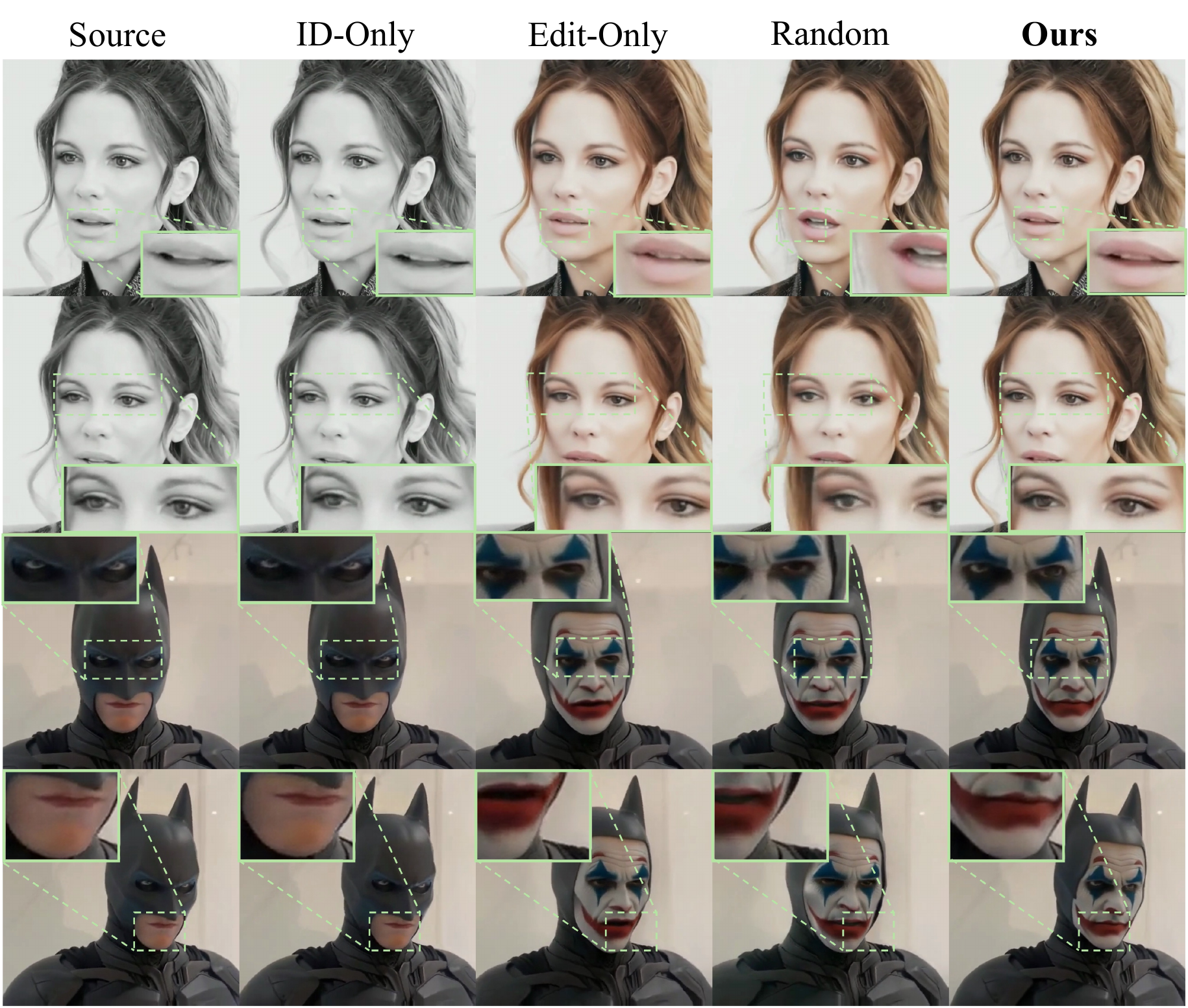}
\end{adjustbox}

\vspace{-0.2cm}

\caption{
\textbf{Effect of dataset composition strategy.}  
Qualitative comparison of different dataset composition strategies. Each column shows a distinct training setup: ID-Only (identical pairs), Edit-Only (edited pairs), Random (unfiltered pairs). Our full method is on rightmost column and the source video is on the left.
}
\vspace{-0.3cm}
\label{fig:ablation_baselines}
\end{figure}

\subsection{Ablation Studies}
\label{subsec:ablations}

In \autoref{fig:ablation_leave_one_out}, we present a \textit{leave-one-out} experiment where we test the contribution of each synchronization channel.
This is done by training models on datasets curated without one component (\emph{speech}, \emph{gaze}, \emph{blink}, or \emph{pose}), which we achieve by setting its respective weight to zero in our filtering score (\autoref{subsec:data_generation_and_curation}).
Concretely, in the first example (top two rows), excluding \emph{pose} leads to visible errors in head orientation, while in the second (bottom two rows), removing any channel introduces spatial drift and temporal misalignment in gaze or blinking.
Omitting any channel degrades synchronization, notably in the man's mouth and the woman's eye motion, confirming that all four cues provide complementary motion information.

In \autoref{fig:ablation_baselines} we ablate different dataset compositions and filtering on model performance.  
We evaluate three baselines:  
\textit{ID-Only}, containing identical (unmodified) pairs;  
\textit{Edit-Only}, containing only edited pairs; and  
\textit{Random}, using unfiltered, randomly generated pairs.  
As illustrated in the top two rows, \textit{ID-Only} maintains alignment but fails to preserve the edit, while \textit{Edit-Only} and \textit{Random} introduce drift in facial regions such as the eyes, mouth, and gaze.  
Our full model achieves both precise synchronization and faithful edit preservation, for example keeping the woman’s eye motion aligned and the painted edit on Batman’s mouth stable, confirming the importance of balanced data and synchronization-based filtering.

A formal quantitative analysis in \autoref{tb:ablations_metrics} confirms these qualitative observations. The leave-one-out study shows removing any single motion cue degrades performance.
Notably, removing the `speech` cue causes the largest drop in Speech Corr. (0.72 to 0.53), confirming its necessity for lip-sync.
Removing the `pose` cue degrades Gaze Corr. (0.75 to 0.67), revealing a learned correlation between head pose and eye movement.

The baseline results in \autoref{tb:ablations_metrics} further prove our data strategy is essential.
The `ID-Only` baseline achieves the highest synchronization (e.g., 0.80 Speech Corr.) but completely fails the edit, scoring only 0.05 in Directional CLIP.
Conversely, the `Edit-Only` and `Random` baselines, which lack our filtering and data balance, suffer from poor synchronization and identity drift.
Our full method is the only one to achieve a strong balance across all three core metrics, confirming our design choices.

\begin{table}[t]
\small
\setlength{\tabcolsep}{4pt}
\addtolength{\belowcaptionskip}{-6pt}
\centering
\begin{adjustbox}{width=\linewidth}
\begin{tabular}{l S[table-format=0.2] S[table-format=0.2] S[table-format=0.2] S[table-format=0.2] S[table-format=0.2] S[table-format=0.2] S[table-format=0.2] S[table-format=0.2]}
    \toprule
    Metric & \multicolumn{4}{c}{Ablations (w/o component)} & \multicolumn{3}{c}{Baselines} & {\textbf{Ours}} \\
    \cmidrule(r){2-5} \cmidrule(r){6-8}
           & {Speech} & {Gaze} & {Blink} & {Pose} & {Only Edit} & {Only ID} & {Random} & \\
    \midrule
    \multicolumn{9}{l}{\textbf{Synchronization}} \\
    Speech Corr. $\uparrow$ & 0.53 & 0.56 & 0.55 & 0.55 & 0.58 & \firstplace{0.80} & 0.56 & \secondplace{0.72} \\
    Gaze Corr. $\uparrow$   & 0.68 & 0.67 & 0.70 & 0.67 & 0.68 & \firstplace{0.83} & 0.68 & \secondplace{0.75} \\
    Blink Corr. $\uparrow$  & 0.46 & 0.45 & 0.49 & 0.45 & 0.48 & \firstplace{0.70} & 0.47 & \secondplace{0.55} \\
    Pose Corr. $\uparrow$   & 0.47 & 0.46 & 0.46 & 0.46 & 0.47 & \firstplace{0.66} & 0.46 & \secondplace{0.55} \\
    \midrule
    \multicolumn{9}{l}{\textbf{Edit Fidelity}} \\
    Directional CLIP $\uparrow$ & 0.55 & \firstplace{0.57} & 0.56 & 0.56 & \firstplace{0.57} & 0.05 & 0.55 & \firstplace{0.57} \\
    Directional CLIP (text-dual) $\uparrow$ & 0.20 & 0.20 & 0.20 & 0.20 & \firstplace{0.21} & 0.01 & 0.20 & \firstplace{0.21} \\
    CLIP-Text Align. $\uparrow$ & \firstplace{0.33} & \firstplace{0.33} & \firstplace{0.33} & \firstplace{0.33} & \firstplace{0.33} & 0.31 & \firstplace{0.33} & \firstplace{0.33} \\
    \midrule
    \multicolumn{9}{l}{\textbf{Identity Preservation}} \\
    ArcFace Sim. $\uparrow$ & 0.72 & 0.72 & 0.72 & 0.72 & \secondplace{0.73} & 0.70 & 0.72 & \firstplace{0.75} \\
    \bottomrule
\end{tabular}
\end{adjustbox}
\vspace{0.0cm}
\caption{\textbf{Impact of Data Curation Strategy.}
We compare our full method ('Ours') against two ablation categories: models trained without specific motion cues (w/o component) and models trained on alternative dataset compositions (Baselines) .
}
\label{tb:ablations_metrics}
\end{table}
\vspace{-0.1cm}

\subsection{Application - Expression Modification}
\label{subsec:expression_change}

\begin{figure}[t]
\centering
\includegraphics[width=\linewidth]{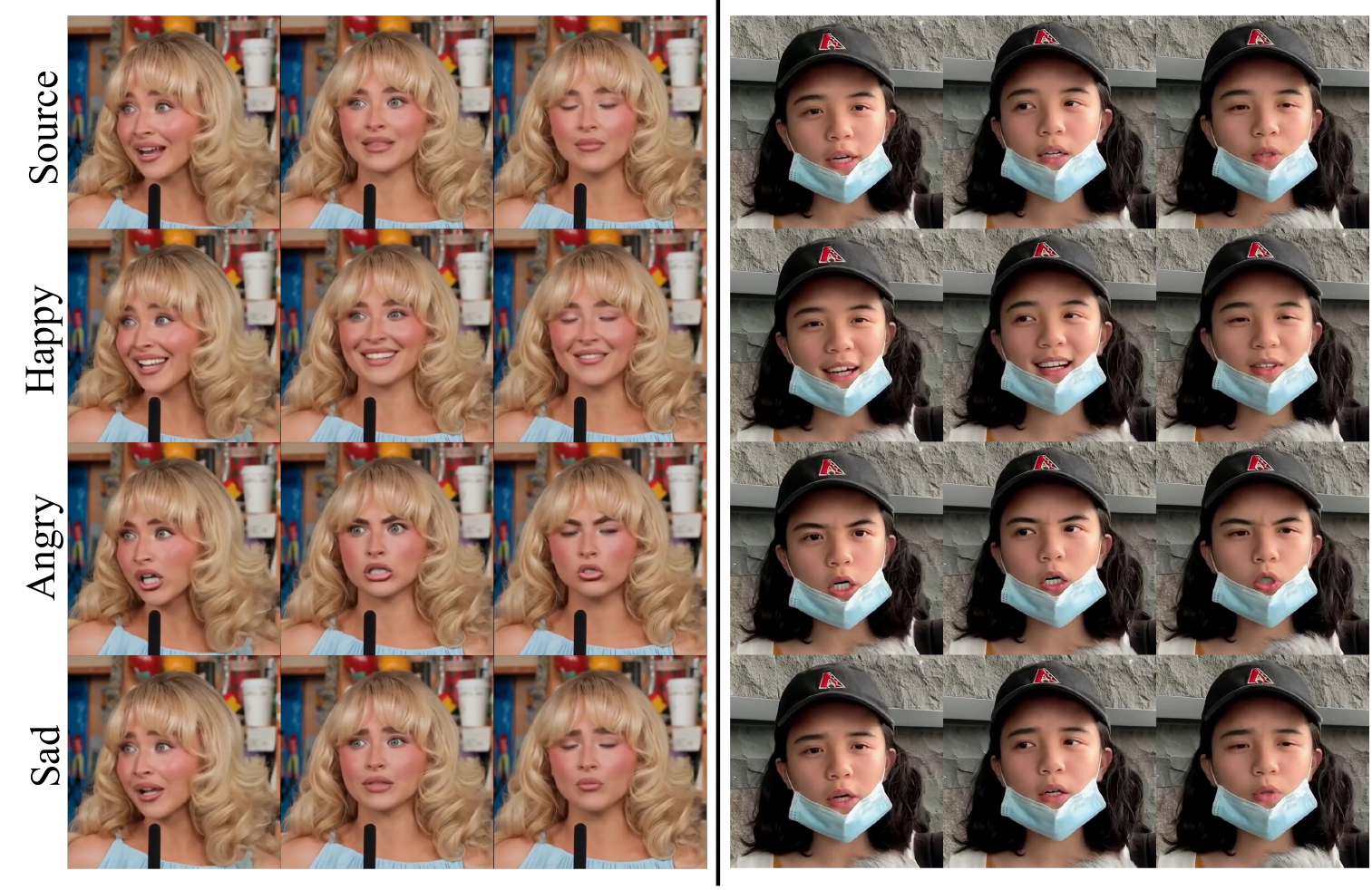}
\caption{\textbf{Expression editing.}
Sync-LoRA performs expression editing (\textit{Happy}, \textit{Angry}, \textit{Sad}) while keeping motion synchronized and geometry consistent, even under occlusions.}
\vspace{-0.2cm}
\label{fig:expressions}
\end{figure}

Demonstrating its capability beyond spatial and appearance modifications, Sync-LoRA also performs expression editing by propagating facial emotion changes while maintaining the underlying articulation.
As illustrated in \autoref{fig:expressions}, our method accurately transfers expressions such as happiness, anger, or sadness to the same source motion, even in the presence of partial occlusions. 
To train an expression-specific Sync-LoRA, we employ LivePortrait~\cite{guo2024liveportrait} to synthesize multiple expressive videos (e.g., happy, angry) from a single neutral reference frame. 
While LivePortrait achieves impressive real-time reenactment, its warping-based synthesis often struggles under extreme head rotations or when occlusions, such as microphones, masks, or hands, partially obscure the face.
In these challenging cases, Sync-LoRA produces more stable and photorealistic results, maintaining structural consistency and accurate synchronization throughout the sequence.
This demonstrates the robustness of our diffusion-based in-context formulation, which models temporal coherence directly in the latent space rather than relying on geometric warping. Comprehensive comparisons highlighting these robustness gains are presented in the supplementary materials.

\section{Conclusions, Limitations and Future work}

Sync-LoRA introduces an in-context LoRA framework for synchronized portrait video editing. By conditioning generation on the original video and its edited first frame, the model learns to propagate local appearance modifications while preserving precise temporal alignment and subject identity. Trained on automatically generated and synchronization-filtered video pairs, Sync-LoRA balances edit fidelity, motion correspondence, and visual realism across diverse subjects and edit types including background manipulation, object insertion, expression modification.

While our method performs robustly across a wide range of scenarios, some limitations remain. The quality of the applied edits may degrade in sequences with fast or large-scale motion, where maintaining fine spatial coherence becomes challenging. Our approach also assumes that the edited first frame is geometrically aligned with the source; significant misalignments can lead to temporal drift or local artifacts during propagation. We provide visual examples of these failure cases in the supplementary material. Resolving these issues remains future work. In addition, since Sync-LoRA builds upon a specific base video diffusion model, future versions could benefit from stronger architectures with enhanced temporal reasoning and multi-view consistency.

Overall, Sync-LoRA establishes a new paradigm for controllable and temporally faithful video editing. This level of frame-accurate synchronization is especially crucial for personalized talking-head applications, where faithfulness to the original action, speech, and performance is essential. Looking ahead, extending this in-context formulation to emerging multi-modal video models that jointly reason over video and audio signals presents an exciting and challenging direction for future research.
\section*{Acknowledgments}
We thank Sigal Raab, Ellie Arar, Nadav Magar, and Mickey Finkelson for their valuable feedback and insightful reviews of this work. This research was supported in part by the Israel Science Foundation (grants no. 2492/20 and 1473/24), Len Blavatnik and the Blavatnik family foundation.
{
    \small
    \bibliographystyle{ieeenat_fullname}
    \bibliography{main}
}

\clearpage
\setcounter{page}{1}


\twocolumn[{%
    \centering
    \Large \textbf{In-Context Sync-LoRA for Portrait Video Editing} \\[0.5em]
    \large Supplementary Material \\[1em]
}]
\def\algorithmautorefname{Algorithm}

\section{Additional Implementation Details}
\label{sec:impl_details_supp}

\subsection{Reference and Target Stream Differentiation}
We implement in-context conditioning by spatially concatenating the source and target streams along the sequence dimension, and differentiating them via per-token timestep conditioning while preserving spatial correspondence. \autoref{alg:stream_diff} details the masking strategy that blocks gradients through the reference stream so it acts as a frozen context while gradients flow only through the target stream.

\begin{algorithm*}[t]
\caption{Pseudocode for Reference and Target Stream Differentiation}
\label{alg:stream_diff}
\begin{lstlisting}[language=Python]
def training_step(source_video, edited_video, diffusion_t):
    """
    source_video: Clean latent frames from the reference video
    edited_video: Noisy latent frames (at time t) from the target video
    diffusion_t: The noise level (timestep) for the target video
    """

    # 1. Sequence Concatenation
    # We concatenate along the sequence dimension (frames).
    # This allows joint attention between source and target.
    model_input = torch.cat([source_video, edited_video], dim=2)

    # 2. Per-Token Timestep Conditioning
    # Source tokens get t=0 (signaling "clean context").
    # Target tokens get t=diffusion_t (signaling "denoise me").
    # These timesteps generate distinct AdaLayerNorm parameters.
    t_source = torch.zeros(source_video.shape[0]) 
    t_target = diffusion_t
    model_timesteps = torch.cat([t_source, t_target], dim=0)

    # 3. Shared Positional Embeddings (RoPE)
    # Both streams use identical grid coordinates (t, h, w).
    # A token at (Frame 1, x, y) in Source has the exact same 
    # positional embedding as (Frame 1, x, y) in Target.
    rope_source = get_3d_rope(source_video.shape)
    rope_target = get_3d_rope(edited_video.shape) # Identical to source
    model_rope = torch.cat([rope_source, rope_target], dim=1)

    # 4. Forward Pass
    # The model uses the shared RoPE to find spatial correspondences
    # and the split timesteps to apply different normalization.
    velocity_pred = model(
        x=model_input, 
        timesteps=model_timesteps, 
        pos_embed=model_rope
    )

    # 5. Loss Masking
    # We only compute loss on the target (edited) half of the sequence.
    pred_source, pred_target = torch.chunk(velocity_pred, 2, dim=2)
    loss = F.mse_loss(pred_target, target_velocity_ground_truth)
    
    return loss
\end{lstlisting}
\end{algorithm*}

\subsection{Hyperparameters}
\label{sec:hyperparams}
We provide a detailed summary of the hyperparameters used for training Sync-LoRA in \autoref{tab:hyperparameters}.

\begin{table}[t]
    \centering
    \small
    \begin{tabular}{lc}
        \toprule
        \textbf{Hyperparameter} & \textbf{Value} \\
        \midrule
        Base Model & LTX-Video \cite{hacohen2024ltxvideorealtimevideolatent} \\
        LoRA Rank & 128 \\
        Optimizer & AdamW \\
        Learning Rate & $2 \times 10^{-4}$ \\
        Batch Size & 1  \\
        Total Training Steps & 5,000 \\ 
        Resolution & $512 \times 512$ \\
        Frames per Clip & 81 \\
        Training Hardware & 1 $\times$ NVIDIA A6000 (48GB) \\
        \bottomrule
    \end{tabular}
    \caption{\textbf{Training Hyperparameters.} Summary of the configuration used for fine-tuning Sync-LoRA.}
    \label{tab:hyperparameters}
\end{table}
    \vspace{-0.5cm}

\section{Data Curation Pipeline}
\label{sec:data_curation_supp}

\begin{figure*}[t]
    \centering
    \includegraphics[width=\linewidth]{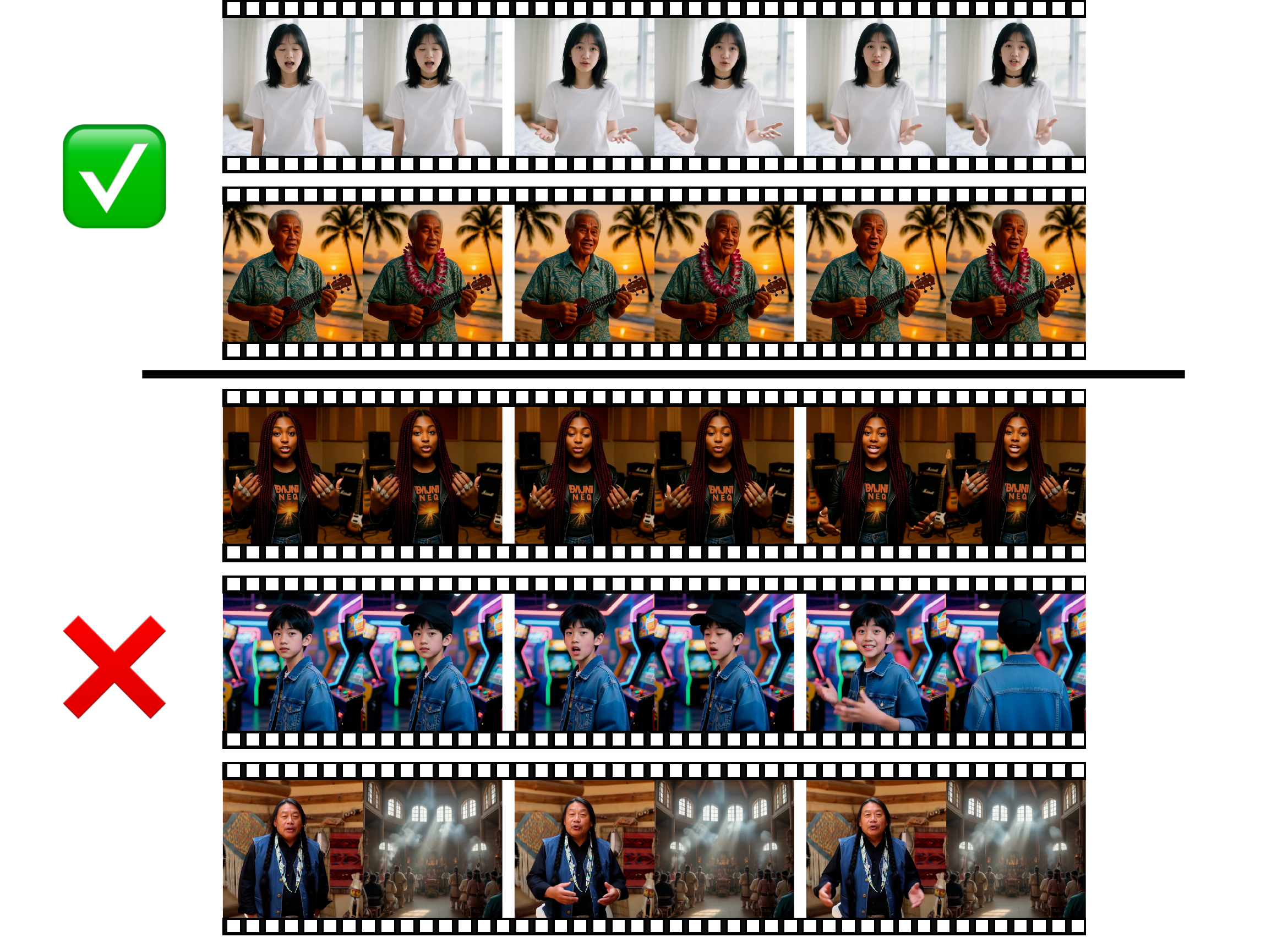}
\caption{\textbf{Data Curation Filtering.} Examples from our automatic filtering pipeline. (Top) \textbf{Accepted} clips exhibit high synchronization scores, indicating precise temporal alignment and stable motion. (Bottom) \textbf{Rejected} clips receive low synchronization scores due to temporal or spatial misalignments and motion drift.}
    \label{fig:filter_examples}
\end{figure*}

This section details the synchronization-based filtering process used to curate our training dataset.
\autoref{fig:filter_examples} illustrates this curation process, showing representative examples of generated pairs that were accepted or rejected based on their synchronization scores.

\subsection{Synchronization Scoring Metrics}
Our filtering pipeline relies on four motion channels derived from MediaPipe \cite{lugaresi2019mediapipeframeworkbuildingperception} landmarks. Let $\mathbf{p}_i \in \mathbb{R}^2$ denote the 2D coordinates of landmark index $i$. For each paired video (source and edited), we extract per-frame scalar signals for each channel as follows:

\begin{itemize}
    \item \textbf{Speech (40\% weight).} We compute the Mouth Aspect Ratio (MAR) as the ratio of the average vertical lip separation to the mouth width. Using the standard MediaPipe mesh indices, we obtain:
    \begin{equation}
        \text{MAR} = \frac{\frac{1}{4}\sum_{j=1}^{4} \|\mathbf{p}_{u_j} - \mathbf{p}_{d_j}\|_2}{\|\mathbf{p}_{61} - \mathbf{p}_{291}\|_2}
    \end{equation}
    where the vertical pairs $(u_j, d_j)$ are $\{(13, 14), (82, 87), (312, 317), (0, 17)\}$, and indices $61, 291$ represent the mouth corners.

    \item \textbf{Gaze (30\% weight).} We calculate the normalized gaze vector $\mathbf{g} = [g_x, g_y]$ representing the iris center relative to the eye bounding box. For the iris center $\mathbf{p}_{\text{iris}}$ and eye boundary landmarks (inner $x_{\text{in}}$, outer $x_{\text{out}}$, upper $y_{\text{up}}$, lower $y_{\text{dn}}$), the normalized coordinates are:
    \begin{equation}
        g_x = 2 \cdot \frac{\mathbf{p}_{\text{iris}, x} - x_{\text{in}}}{x_{\text{out}} - x_{\text{in}}} - 1, \quad
        g_y = 2 \cdot \frac{\mathbf{p}_{\text{iris}, y} - y_{\text{up}}}{y_{\text{dn}} - y_{\text{up}}} - 1
    \end{equation}
    We compute this for both eyes and average the motion vectors.

    \item \textbf{Blink (15\% weight).} We utilize a simplified Eye Aspect Ratio (EAR) based on the ratio of vertical to horizontal eye landmarks. The signal is averaged across both eyes and negated so that blink events appear as positive peaks:
    \begin{equation}
        \text{Signal}_{\text{blink}} = - \frac{1}{2} \left( \frac{\|\mathbf{p}_{159} - \mathbf{p}_{145}\|}{\|\mathbf{p}_{33} - \mathbf{p}_{133}\|} + \frac{\|\mathbf{p}_{386} - \mathbf{p}_{374}\|}{\|\mathbf{p}_{263} - \mathbf{p}_{362}\|} \right)
    \end{equation}

    \item \textbf{Pose (15\% weight).} We track six upper-body geometric features to capture torso dynamics:
    \begin{enumerate}
        \item \textbf{Shoulder Orientation:} Angle of the vector $\mathbf{p}_{12} - \mathbf{p}_{11}$.
        \item \textbf{Torso Inclination:} Angle of the vector connecting the shoulder midpoint to the hip midpoint ($\mathbf{p}_{23}, \mathbf{p}_{24}$).
        \item \textbf{Elbow Angles (Left/Right):} Angles formed by the Shoulder-Elbow-Wrist triplets (11-13-15 and 12-14-16).
        \item \textbf{Relative Wrist Heights (Left/Right):} The vertical displacement between shoulder and wrist, defined as $h = y_{\text{shoulder}} - y_{\text{wrist}}$.
    \end{enumerate}
\end{itemize}

\noindent
\textbf{Signal Processing:} To ensure robustness against detection noise, all raw signals undergo the following processing steps:
\begin{enumerate}
    \item \textbf{Interpolation:} Missing detections (NaNs) are filled via linear interpolation.
    \item \textbf{Smoothing:} We apply a Savitzky-Golay \cite{Savitzky–Golay} filter with a window length of $W=9$ and polynomial order $K=2$.
    \item \textbf{Normalization:} Signals are z-score normalized, $z = \frac{x - \mu}{\sigma + \epsilon}$ , with $\epsilon=10^{-6}$.
\end{enumerate}

\noindent
\textbf{Final Score:} The final synchronization score is computed as the weighted sum of the Pearson correlation coefficients (at zero temporal lag) between the processed signals of the source and edited videos.

\section{Evaluation Benchmark and Metrics}
\label{sec:eval_benchmark_supp}

\subsection{Benchmark Curation}
\begin{figure}[t]
    \centering
    \includegraphics[width=\linewidth]{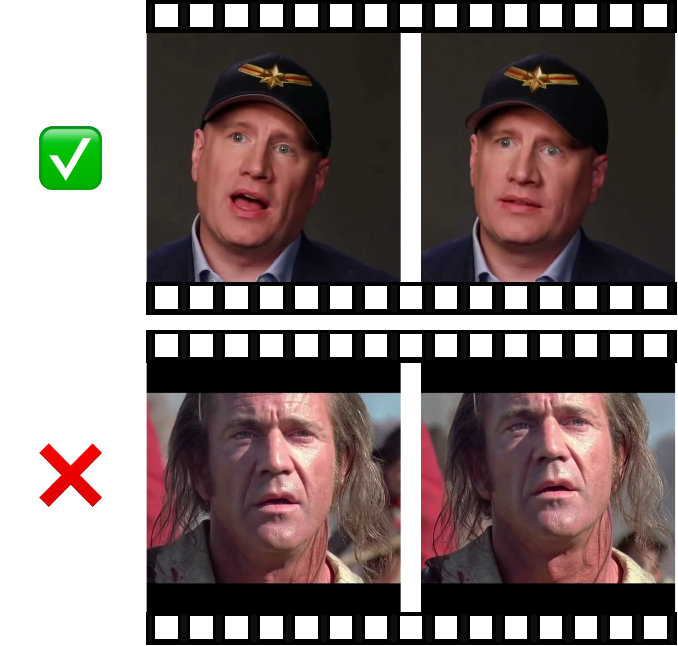}
    \caption{\textbf{Benchmark Examples.} 
examples}

    \label{fig:sup_benchmark_examples}
\end{figure}

As mentioned in Section 4 of the main paper, our evaluation benchmark was strictly curated to ensure high-quality assessment. We applied rigorous criteria, retaining only clips with consistent lighting and high sharpness, while excluding those with jump cuts, subtitles, or significant motion blur. \autoref{fig:sup_benchmark_examples} illustrates a typical rejection case, such as a clip with noticeable letterboxing.

\subsection{Quantitative Evaluation Metric Definitions}
\label{sec:metrics_definitions}
This subsection provides the full mathematical definitions of the metrics reported in Subsection 4.2 of the main paper.
All notations follow those used in the main paper.
Given source frames \(I^{\mathrm{src}}_t\), edited frames \(I^{\mathrm{edit}}_t\), and an edited keyframe \(I^{\mathrm{key}}\), the following metrics are used for quantitative evaluation.

\paragraph{Directional CLIP Score.}
This metric measures how consistently the per-frame visual transformation aligns with the intended edit direction in CLIP’s embedding space.
We report two variants: an \textbf{image-based} version and a \textbf{text-dual} version. The CLIP model used is `ViT-B-32'.

\textit{Image-based Direction.}
The global edit direction in CLIP image space is defined as:
\[
\mathbf{d}_{\text{img}} =
\frac{E_{\text{img}}(I^{\mathrm{key}}) - E_{\text{img}}(I^{\mathrm{src}}_0)}
{\|E_{\text{img}}(I^{\mathrm{key}}) - E_{\text{img}}(I^{\mathrm{src}}_0)\|}.
\]
For each frame \(t\), we compute its local transformation direction:
\[
\mathbf{V}_t =
\frac{E_{\text{img}}(I^{\mathrm{edit}}_t) - E_{\text{img}}(I^{\mathrm{src}}_t)}
{\|E_{\text{img}}(I^{\mathrm{edit}}_t) - E_{\text{img}}(I^{\mathrm{src}}_t)\|}.
\]
The Directional CLIP score is their mean cosine similarity:
\[
\text{DirectionalCLIP}_{\text{img}}
= \frac{1}{T} \sum_{t=1}^{T} \mathbf{V}_t \cdot \mathbf{d}_{\text{img}}.
\]

\textit{Text-dual Direction.}
For methods guided by text rather than edited keyframes, the semantic direction is obtained from CLIP’s text encoder:
\[
\mathbf{d}_{\text{text}} =
\frac{E_{\text{text}}(t_{\text{target}}) - E_{\text{text}}(t_{\text{source}})}
{\|E_{\text{text}}(t_{\text{target}}) - E_{\text{text}}(t_{\text{source}})\|}.
\]
The local per-frame directions \(\mathbf{V}_t\) are identical to the image-based case, yielding:
\[
\text{DirectionalCLIP}_{\text{text-dual}}
= \frac{1}{T} \sum_{t=1}^{T} \mathbf{V}_t \cdot \mathbf{d}_{\text{text}}.
\]
Higher values indicate stronger alignment between frame-wise visual changes and the intended edit.

\paragraph{CLIP-Text Alignment Score.}
This metric evaluates the semantic consistency of the edited video with the target description.
Given normalized text embedding \(\mathbf{t} = E_{\text{text}}(\cdot)\) and per-frame image embeddings \(E_{\text{img}}(I^{\mathrm{edit}}_t)\):
\[
\text{CLIPTextAlign} =
\frac{1}{T} \sum_{t=1}^{T} E_{\text{img}}(I^{\mathrm{edit}}_t) \cdot \mathbf{t}
\]

\paragraph{ArcFace Similarity.}
To quantify identity preservation, we compute the mean cosine similarity between ArcFace~\cite{Deng_2022_ArcFace} embeddings of each frame and the edited keyframe:
\[
\text{ArcFaceSim} =
\frac{1}{T} \sum_{t=1}^{T}
E_{\text{face}}(I^{\mathrm{edit}}_t) \cdot E_{\text{face}}(I^{\mathrm{key}})
\]
Higher values correspond to stronger identity consistency across frames.

\section{User Study}
\label{sec:user_study_supp}

We conducted a user study to evaluate Sync-LoRA based on four distinct criteria: (1) \textbf{Edit Fidelity} (adherence to the visual instruction), (2) \textbf{Synchronization} (preservation of timing and movement), (3) \textbf{Identity Preservation} (consistency with the subject's original identity), and (4) \textbf{Overall Preference}. To this end, we performed a blinded, randomized Two-Alternative Forced Choice (2AFC) comparison between our method and four state-of-the-art baselines: VACE, LucyEdit, AnyV2V, and FlowEdit. We curated a robust pool of 36 distinct video comparison pairs distributed across three unique evaluation forms. The study involved 23 independent participants, where each completed 12 comparisons (3 per baseline), yielding a total of 69 pairwise judgments for each competitor method.

\begin{figure}[t!]
\small
\setlength{\tabcolsep}{4pt}
\centering
\includegraphics[width=0.45\textwidth]{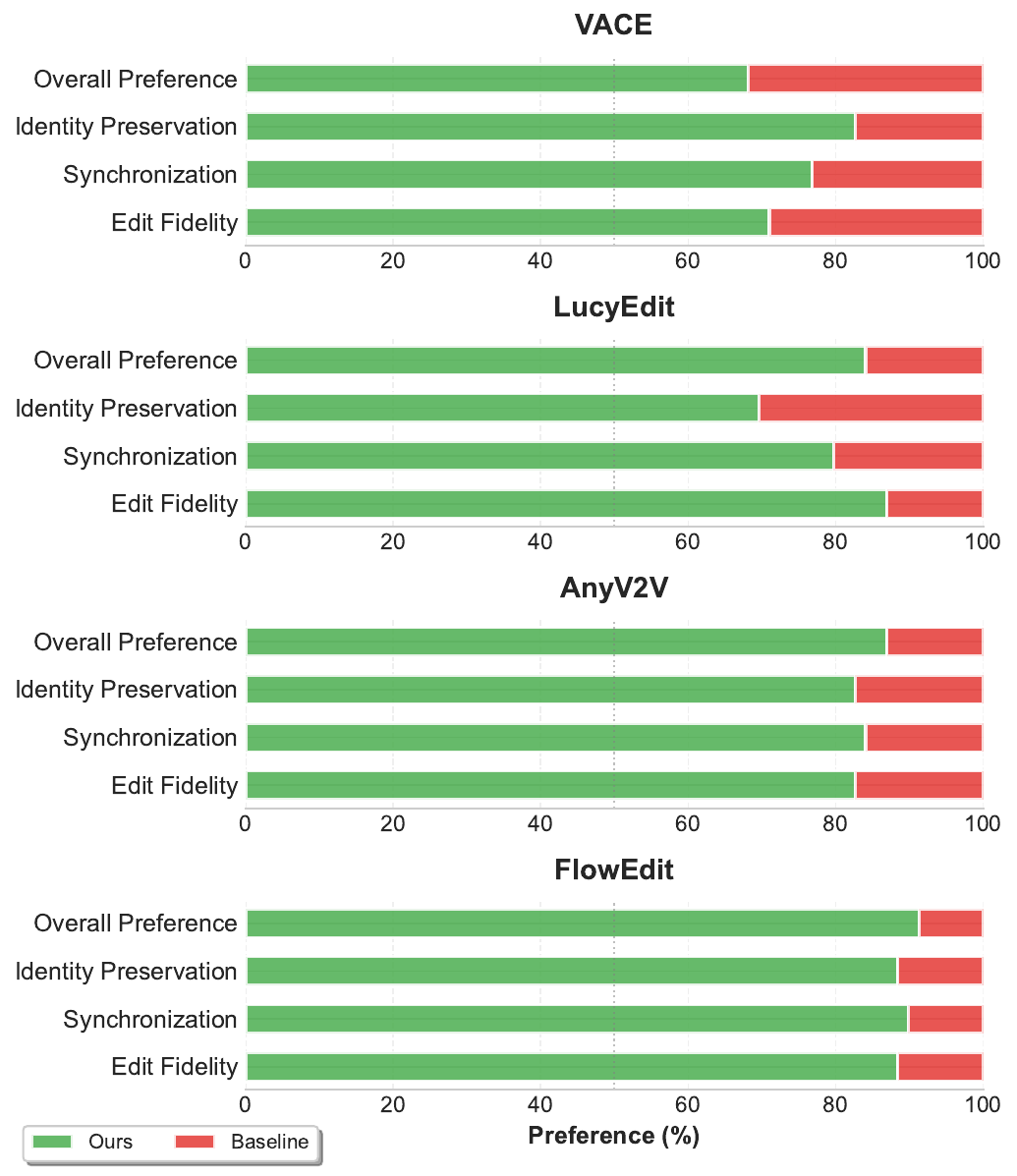} \\[-0.4cm]
\vspace{0.5cm}
\caption{\textbf{User Study Results.} Pairwise preference rates for Sync-LoRA (Ours) against four baselines. Our method is strongly preferred across all four criteria: Edit Fidelity, Synchronization, Identity Preservation, and Overall Preference.}
\vspace{-0.2cm}
\label{fig:user_study}
\end{figure}

The complete results are presented in \autoref{fig:user_study}, where we report the preference rates for Sync-LoRA against each baseline. As shown, users strongly favored our approach by a significant margin across all metrics. Notably, our method achieved the highest gains in Identity Preservation and Synchronization, confirming our core contribution of maintaining temporal coherence and subject identity while faithfully applying the intended edits.

\section{Supplementary Videos}
\label{sec:video_supp}

Qualitative evaluation of video editing requires assessing motion stability and temporal coherence, which static frames cannot fully convey. Therefore, we strongly encourage readers to view the results in the \emph{attached supplementary webpage}. This interactive file features comprehensive side-by-side comparisons against all baselines (VACE, LucyEdit, FlowEdit, AnyV2V), detailed ablation studies verifying our synchronization cues, and extensive examples of expression editing and failure cases.

\section{Expression Modification}
\label{sec:expression_supp}

This section provides comparisons for the expression editing application mentioned in Subsection 4.4 of the main paper. Specifically, we compare our results against LivePortrait \cite{guo2024liveportrait} to demonstrate how our diffusion-based approach resolves emergent issues typical of warping-based modules.

\begin{figure}[t]
\centering
\includegraphics[width=\linewidth]{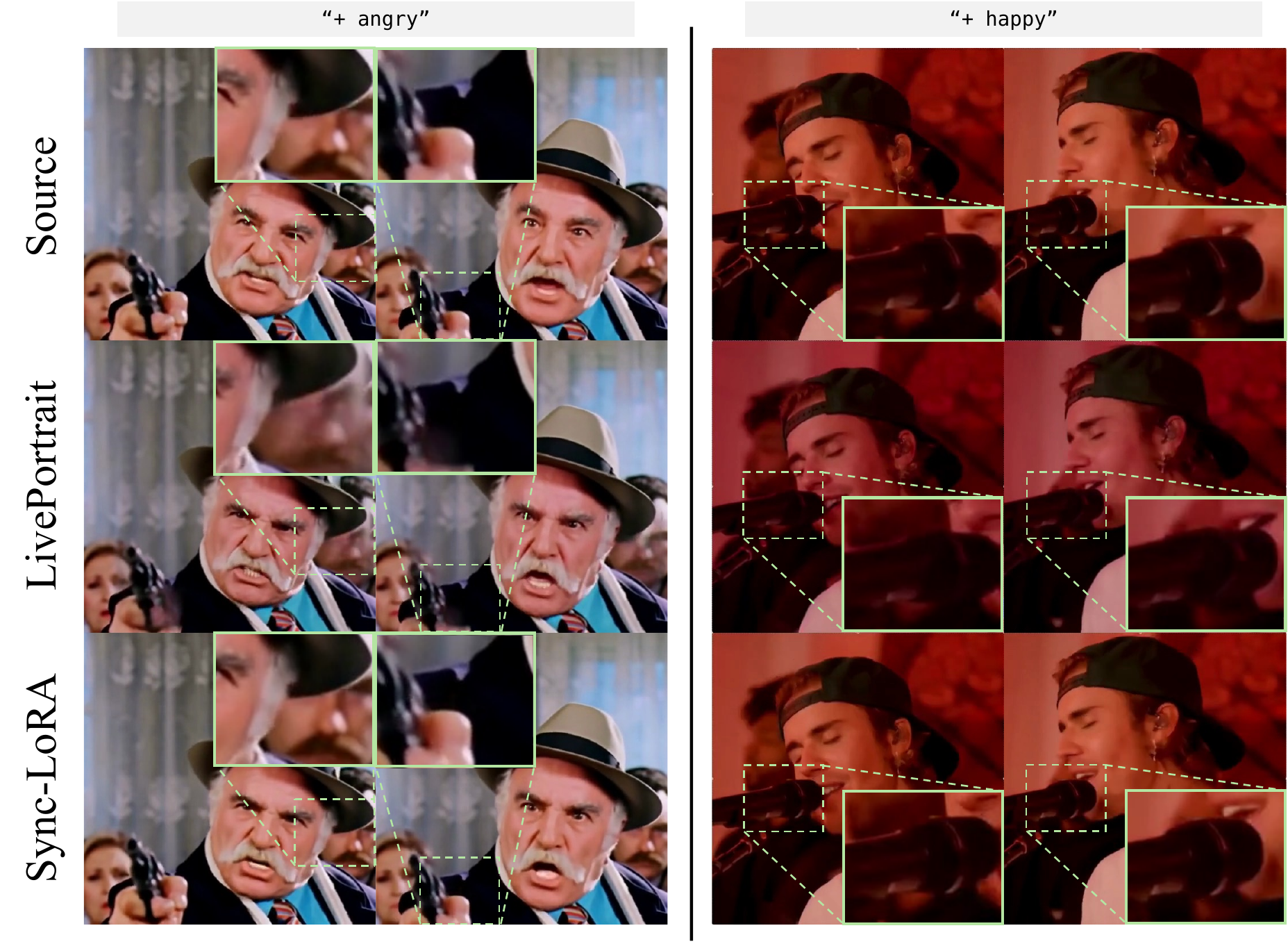}
\caption{\textbf{Expression editing robustness.} Comparison between LivePortrait \cite{guo2024liveportrait} and Sync-LoRA (Ours).}\vspace{-0.2cm}
\label{fig:sup_expressions}
\end{figure}

\textbf{Robustness to Occlusion and Rotation.} LivePortrait employs warping fields to transfer expressions from a driving video. As shown in \autoref{fig:sup_expressions}, this approach often produces artifacts when the face is partially occluded by objects or when background elements are in close proximity. For instance, in the left example, the warping field erroneously deforms the background person's face and blurs the hand holding the pistol. Similarly, in the right example, the rigid structure of the microphone is unnaturally bent near the subject's jawline. In contrast, Sync-LoRA treats expression editing as an in-context generation task. By leveraging the strong generative prior of the diffusion model, our method synthesizes geometrically plausible pixels even in occluded regions, maintaining structural integrity and photorealism while accurately transferring the target expression.

\section{Limitations and Failure Cases}
\label{sec:failures_supp}

\begin{figure}[t]
\centering
\includegraphics[width=\linewidth]{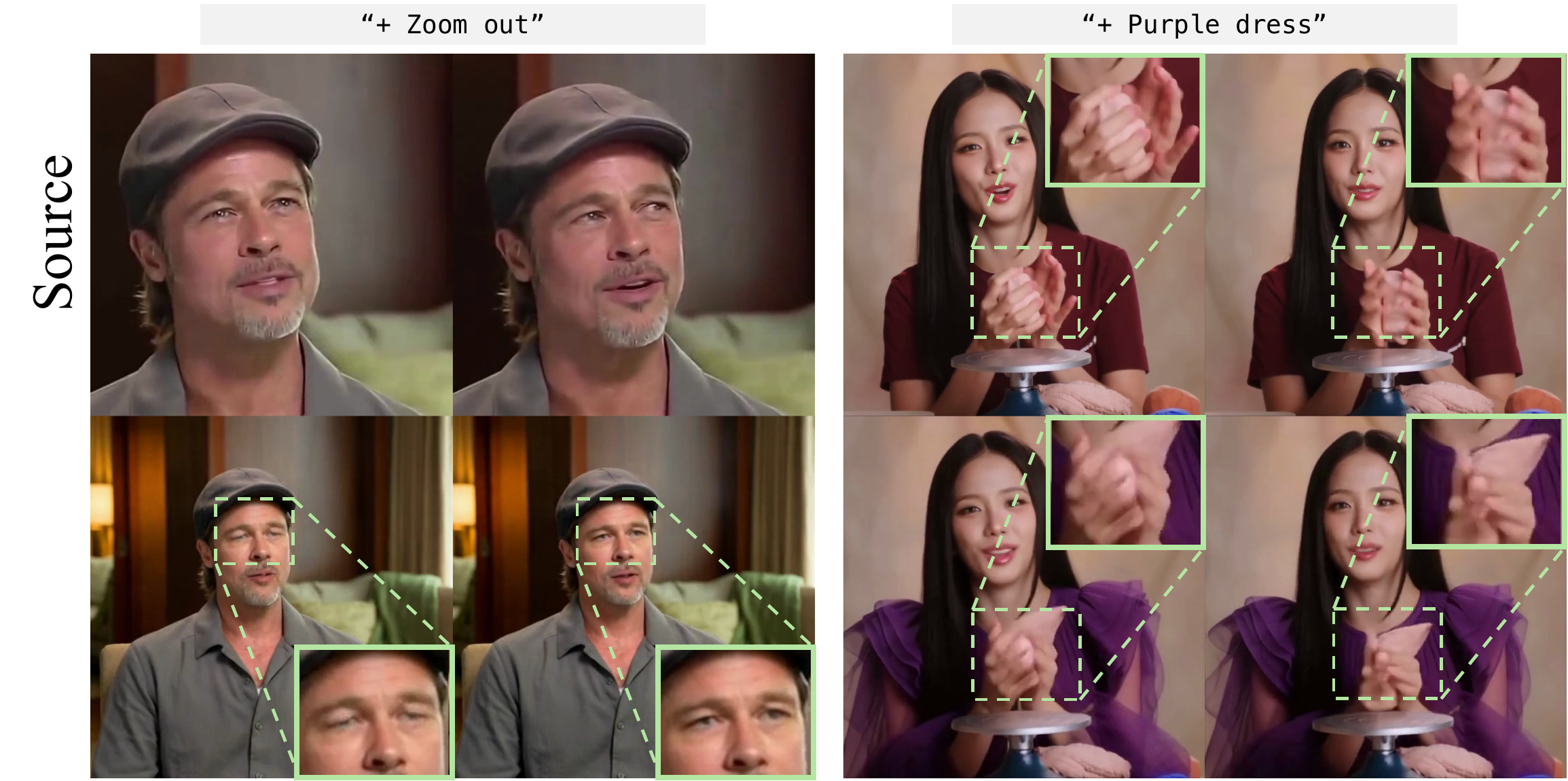}
\caption{\textbf{Limitations of Sync-LoRA.} The figure illustrates two primary failure modes. (Left) Spatial and synchronization degradation during non-aligned geometric edits (zoom-out), resulting in blurred facial features and temporal drift. (Right) Loss of detail and warping artifacts on fast-moving, complex regions (hands) during complex appearance modifications.}
\vspace{-0.2cm}
\label{fig:sup_limitations}
\end{figure}

As discussed in Section 5 of the main paper, our method exhibits specific failure modes. We categorize these into two primary types, illustrated in \autoref{fig:sup_limitations}.

\textbf{Geometric Misalignment.} Our method conditions the generation on both the edited first frame and the source video context. If the edited frame structurally contradicts the spatial information inherent in the source video (e.g., a zoom-out as shown in \autoref{fig:sup_limitations}), the model attempts to reconcile two conflicting spatial signals. This often compromises temporal alignment, leading to noticeable synchronization issues that propagate through the sequence.

\textbf{Rapid Motion Degradation.} In sequences involving very fast or large-scale motion (such as rapid hand movements, dancing, or camera pans), the optical flow guidance from the source video can become ambiguous, resulting in blurred textures or a loss of temporal coherence. We anticipate that this limitation could be mitigated by future base models with enhanced temporal reasoning capabilities.

\end{document}